\documentclass[conference, letterpaper, 10pt]{IEEEtran}
\IEEEoverridecommandlockouts
\usepackage{caption}
\usepackage{graphicx}
\usepackage{amsmath, amssymb}
\usepackage{bbm}
\usepackage{cite}
\usepackage{mathtools}
\usepackage{algorithm}
\usepackage{algorithmic}
\usepackage{booktabs}
\usepackage{caption}
\usepackage{subcaption}
\usepackage{bbold}
\usepackage{dsfont}
\usepackage{amsfonts}
\usepackage{tikz}
\usepackage{authblk}
\usepackage{tabularx}
\usepackage{xcolor}
\usepackage{enumitem}

\usetikzlibrary{arrows,automata,positioning, shapes.misc}

\usepackage{xcolor}

\definecolor{questionblue}{RGB}{46, 85, 115}
\definecolor{questionbg}{RGB}{232, 240, 247}
\definecolor{graphorange}{RGB}{180, 75, 40}
\definecolor{graphbg}{RGB}{250, 230, 220}
\definecolor{tradpurple}{RGB}{120, 32, 110}
\definecolor{tradbg}{RGB}{248, 231, 246}

\newsavebox{\labeledboxcontent}

\newenvironment{labeledbox}[3]{%
  \def\tabcolor{#1}%
  \def\bgcolor{#2}%
  \def\tablabel{#3}%
  \begin{lrbox}{\labeledboxcontent}%
  \begin{minipage}{\dimexpr\linewidth-20pt\relax}%
  \normalsize
}{%
  \end{minipage}%
  \end{lrbox}%
  \noindent
  \begin{tikzpicture}
    \node[fill=\bgcolor, inner sep=4pt] (box) {\usebox{\labeledboxcontent}};
    \node[fill=\tabcolor, text=white, font=\scriptsize\bfseries,
          inner xsep=10pt, inner ysep=4pt,
          anchor=south west] at ([xshift=8pt, yshift=-2pt]box.north west) {\tablabel};
  \end{tikzpicture}%
}
\makeatletter  

\title{BEACON-SP: Ontology-Grounded GraphRAG Framework for Clinical Suicide Risk Assessment}

\author{
\normalsize Kemal Davaslioglu\textsuperscript{1},
Nathan Conger\textsuperscript{1},
Sastry Kompella\textsuperscript{1},
Yalin E. Sagduyu\textsuperscript{1},
and Nathaniel D. Bastian\textsuperscript{2}
\\
\small \textsuperscript{1}Nexcepta, Gaithersburg, MD, USA \\
\small \textsuperscript{2}United States Military Academy, West Point, NY, USA

\thanks{This research was, in part, funded by the U.S. Government. The views and conclusions contained in this document are those of the authors and should not be interpreted as representing the official policies, either expressed or implied, of the U.S. Government.}
}

\date{}
\begin{document}

\maketitle
\thispagestyle{empty}

\begin{abstract}
We present BEACON-SP, an ontology-grounded Graph Retrieval-Augmented Generation (GraphRAG) framework for clinician-facing decision support in behavioral health settings such as suicide prevention, where effective assessment requires integrating heterogeneous clinical, behavioral, social, and temporal evidence. BEACON-SP combines patient knowledge graphs with ontology-guided retrieval to support multi-hop reasoning across diagnoses, medications, risk and protective factors, life events, and temporal relationships. The framework is enabled by a comprehensive suicide prevention ontology that integrates the Three-Step Theory, the Integrated Motivational-Volitional Model, and the Suicide Social Determinants of Health Ontology into a unified representation of patient risk factors. We construct ontology-grounded patient knowledge graphs and evaluate BEACON-SP for clinician-facing question answering. Compared with a vector-based retrieval-augmented generation (RAG) baseline on a 1,500-query benchmark spanning 15 clinical categories and 100 patients, BEACON-SP improves completeness, clinical relevance, and evidence grounding under a corrected comparative evaluation protocol, with a small gain on factual accuracy. In paired criterion-level comparisons, GraphRAG is preferred in 76.4\% of cases. These results demonstrate the potential of ontology-guided GraphRAG to provide structured, contextualized patient evidence for clinical decision support.

\end{abstract}
\section{Introduction}

The growing complexity of modern healthcare has increased the need for intelligent tools that can integrate heterogeneous clinical, behavioral, social, and temporal information. Despite advances in AI, synthesizing such evidence into actionable clinical assessments remains challenging, particularly in suicide prevention where risk is driven by interacting psychiatric, social, environmental, and temporal factors.

Suicide is one of the leading causes of death in the United States, posing a profound public health challenge. According to the Substance Abuse and Mental Health Services Administration (SAMHSA), more than 14 million people in the United States contemplated or attempted suicide in 2024 \cite{samhsa_nsduh_2024}. Suicide is the second leading cause of death among Veterans under age 45 \cite{USVetReport}. In 2023, Veterans accounted for 6,398 of 47,711 U.S. adult suicides, a rate of 35.2 per 100,000 Veterans \cite{USVetReport}. Risk is driven by psychiatric conditions, including anxiety, depression, PTSD, and substance use disorders, as well as social determinants such as housing, financial security, and healthcare access. Despite this complexity, existing clinical tools do not comprehensively integrate diagnoses, social determinants, and behavioral indicators into a unified risk assessment. Consequently, they provide limited support for timely, personalized intervention and evaluation of veteran suicide risk.

In this paper, we present Better Evaluation and Assessment for Comprehensive  Mental Health Outcomes and Navigation Suicide Prevention (BEACON-SP), an ontology-grounded Graph Retrieval Augmented Generation (GraphRAG) framework for clinician-facing decision support in behavioral health settings. BEACON-SP combines patient-specific knowledge graphs, ontology-guided retrieval, and graph-based evidence expansion to support multi-hop reasoning across diagnoses, medications, risk factors, protective factors, life events, and temporal relationships. The framework is not intended to autonomously assess suicide risk or replace clinician judgment. Instead, it serves as a clinician-assistive system that retrieves and contextualizes patient evidence, identifies relevant risk and protective factors, maps clinical findings to established suicide theories, and supports structured risk formulation. BEACON-SP incorporates a comprehensive suicide prevention ontology and a reproducible pipeline for constructing patient knowledge graphs from Electronic Health Records (EHRs). 

This paper makes the following contributions:

\begin{itemize}
\item We introduce BEACON-SP, an ontology-grounded GraphRAG framework that combines semantic retrieval and graph traversal to support clinician-facing question answering and structured suicide-risk assessment.

\item We design a two-pass GraphRAG retrieval strategy that combines cosine-similarity seed retrieval with breadth-first graph expansion to capture clinically relevant relational and temporal context.

\item We introduce a comprehensive suicide prevention ontology integrating the Three-Step Theory (3ST), the Integrated Motivational-Volitional (IMV) model, and the Suicide SDoH Ontology (Suicide-SDoHO) into a unified patient-level knowledge representation.

\item We present a reproducible pipeline for constructing ontology-grounded patient knowledge graphs from EHRs.

\item We evaluate BEACON-SP GraphRAG against a vector-based RAG baseline across 1,500 clinical queries spanning 15 suicide-prevention tasks and 100 patients.

\item We identify and correct a systematic evaluation bias that underestimates graph-based retrieval systems, and introduce a comparative protocol that evaluates each system against its retrieved evidence.
\end{itemize}

The remainder of this paper is organized as follows. Section~\ref{sec:related_works} reviews related work. Section~\ref{sec:suicide_risk_modeling} summarizes the suicide-risk theories that motivate BEACON-SP. Section~\ref{sec:ontology_framework} presents the ontology and knowledge graph. Section~\ref{sec:rag_graphrag} describes the RAG and ontology-guided GraphRAG workflows. Section~\ref{sec:experiments} presents the benchmark design and evaluation results. Section~\ref{sec:discussion} concludes the paper.

\section{Related Work}\label{sec:related_works}
Recent work has explored RAG, knowledge graphs, and Large Language Model (LLM)-based assistants for healthcare, mental health, and clinical decision support. A GraphRAG that links a corpus of medical literature to MIMIC-IV EHRs is applied in \cite{wu2024medical} to improve evidence-based medical responses from LLMs. A two-stage architecture for clinical risk prediction is developed in \cite{Datta2025ImprovingHR} in which a fine-tuned LLM extracts task-relevant information from clinical notes enriched by knowledge-graph retrieval and a second stage combines this unstructured evidence with structured data to generate final predictions. TrumorGPT, a GraphRAG-based LLM framework is introduced in \cite{hang2025trumorgpt} for fact-checking health-related claims using semantic health knowledge graphs and semantic reasoning. A mental-health LLM is combined with GraphRAG in \cite{guo2024knowledge} for depression detection. LLM-Therapist \cite{shafi2024llmtherapist} combines RAG and function calling in a multimodal healthcare assistant for personalized responses. Our prior work \cite{davasli26} explored contextual RAG for domain-specific knowledge discovery; BEACON-SP extends this approach with ontology-grounded patient graphs for relational and temporal reasoning.

In contrast to these approaches, BEACON-SP organizes patient evidence through an ontology-based schema that unifies 3ST theory, IMV model, and Suicide-SDoHO ontology. The resulting patient knowledge graph explicitly represents psychiatric diagnoses, medications, social determinants, risk factors, protective factors, suicide-theory constructs, and temporal risk windows. GraphRAG then operates over this patient-specific graph to support both single- and multi-hop clinician-facing questions, which provides theory-grounded retrieval context rather than only semantically similar text chunks or generic mental health guidance.

\section{Suicide Risk Modeling}\label{sec:suicide_risk_modeling}

BEACON-SP draws on three complementary frameworks for representing patient-level suicide risk.

\textbf{Three-Step Theory (3ST)} characterizes suicide risk through psychological pain, hopelessness, connectedness, and capacity for suicide~\cite{3st}. Ideation emerges when pain and hopelessness co-occur, its severity depends on whether connectedness offsets pain, and progression to attempt depends on dispositional, acquired, and practical capacity.

\textbf{Integrated Motivational-Volitional (IMV) Model} organizes suicide risk into pre-motivational, motivational, and volitional phases~\cite{IMV}. These phases capture background vulnerability, development of suicidal ideation through defeat and entrapment, and transition from intent to behavior through factors such as means access, impulsivity, exposure to suicide, and fearlessness about death.

\textbf{Suicide SDoH Ontology (Suicide-SDoH)} provides a structured vocabulary for social and contextual contributors to suicide risk, including social context, behavior and lifestyle, economic stability, and suicide-specific concepts~\cite{Suicide_SDOHO}.

Together, these frameworks support a unified patient-level ontology linking clinical observations to theory-grounded risk constructs and temporal context.

%Together, 3ST and IMV define the ideation-to-action pathway, while Suicide-SDoH captures social and environmental contributors. BEACON-SP integrates these frameworks into a unified ontology that can be instantiated as patient-level knowledge graphs, linking clinical observations to theory-grounded risk constructs and temporal context.

\section{Ontology Framework}\label{sec:ontology_framework}

\begin{figure}[t!]
   \centering
   \includegraphics[width=\linewidth]{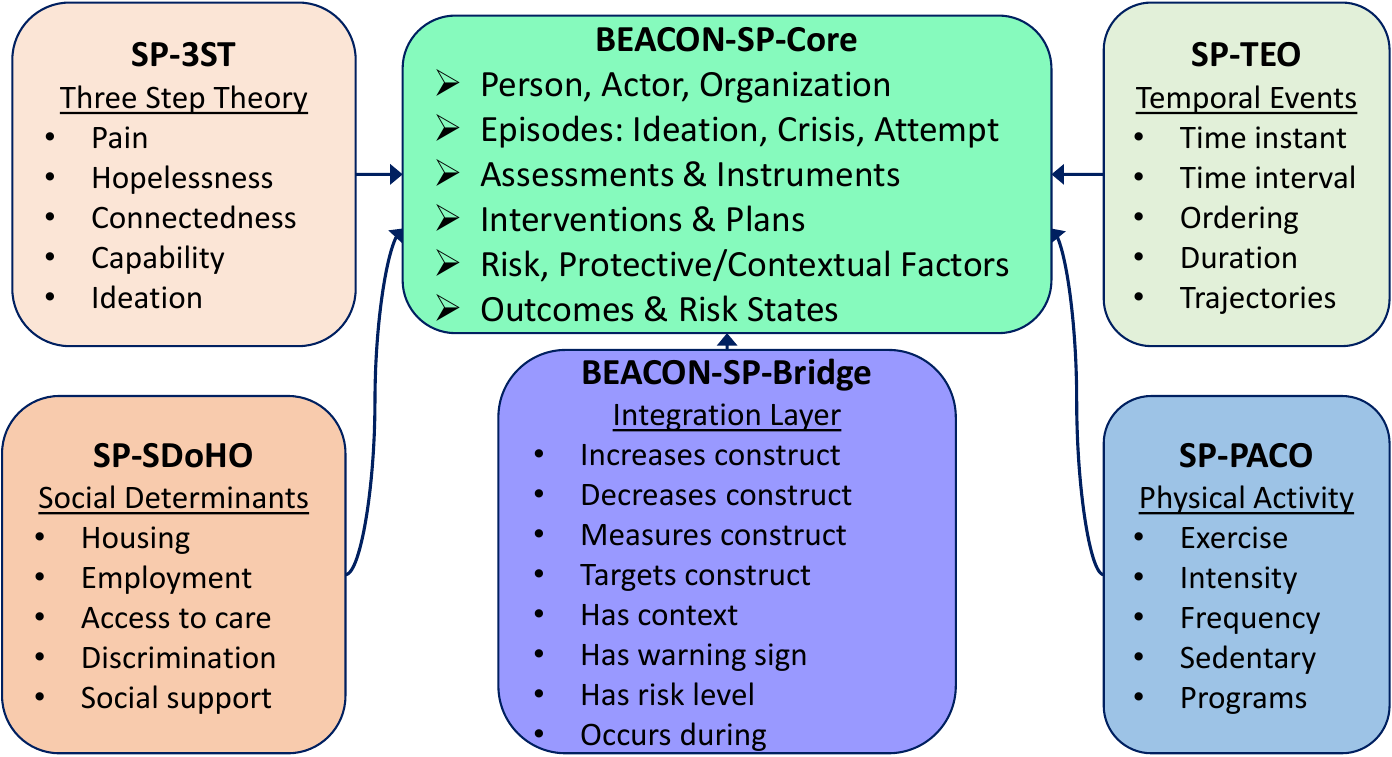}
   \caption{BEACON-SP Ontology components.}
   \label{Fig:BEACON-SP_Flow}
\end{figure}

Figure~\ref{Fig:BEACON-SP_Flow} illustrates the high-level architecture of BEACON-SP which shows how the 3ST, Suicide-SDoH, the Physical Activity Concept Ontology (PACO) \cite{kim2019paco}, and the Time Entity Ontology (TEO) \cite{li2020teo} connect to the BEACON Core, with the BEACON Bridge layer that unifies patient features. The BEACON-SP ontology contains 247 attributes organized across 12 feature domains which cover mental health information, medications, life events, risk and protective factor profiles, and a temporal component that represents processes and events over time. 

The feature domains and their principal additions are as follows: \textbf{Risk factors} are derived directly from the Suicide-SDoH across ten categories (behavioral, demographic, developmental, economic, education, legal, neighborhood, neurobiological, psychological, and social), replacing earlier generalized categories with clinically grounded suicide-specific representations.
\textbf{Protective factors} capture conditions associated with reduced suicide risk, including social connectedness, coping skills, stable employment, means restriction, and effective clinical care. \textbf{Suicide behavior spectrum} represents the progression from passive and active ideation through suicidal intent, planning, and behavior, incorporating constructs from both IMV and 3ST. 
\textbf{Life events} encode triggering clinical and social episodes such as hospitalization, discharge, job loss, eviction, bereavement, and military deployment, as timestamped nodes. \textbf{Medical conditions} include disorders with established associations to elevated suicide risk: chronic pain, traumatic brain injury, sleep disorders, epilepsy, cancer, HIV, and chronic kidney disease. \textbf{Medications} cover psychiatric pharmacotherapy organized by class: antidepressants (e.g., bupropion, mirtazapine etc.), antipsychotics, anxiolytics, mood stabilizers, and ketamine, with interaction effects modeled where multiple agents are co-prescribed. 
\textbf{Psychiatric diagnoses} span major disorder categories from the DSM5-TR with documented suicide associations: mood, psychotic, personality, anxiety, substance use, trauma-related, and neurodevelopmental disorders. 
\textbf{Interventions} include hospitalization, psychotherapy modalities, safety planning, lethal means counseling, crisis intervention, and related crisis-response treatments. \textbf{Assessments and biomarkers} incorporate validated clinical instruments (e.g., PHQ-9, GAD-7, CSSRS, PCL-5, AUDIT, Beck-SSI) alongside biological markers (cortisol, serotonin, C-reactive protein, IL-6) that index biological vulnerability. 
\textbf{Temporal features} represent four time-dependent constructs: state transitions (discrete timestamped clinical or life changes), clinical epochs (treatment or contextual phases such as hospitalization or incarceration), observation series (longitudinal repeated measurements of the same clinical variable), and risk windows (bounded periods of elevated risk triggered by specific events such as post-discharge or medication initiation).

\section{Patient Knowledge Graph and GraphRAG Implementation}\label{sec:rag_graphrag}

In a vector-based RAG pipeline, documents and patient records are first divided into chunks, and each chunk is encoded into a dense vector representation. A user query is encoded using the same embedding model, and the retriever ranks candidate chunks by similarity in the embedding space. In our baseline, this similarity is computed using cosine similarity, $\mathrm{sim}(\mathbf{q},\mathbf{d}_i)= \mathbf{q}^{\top}\mathbf{d}_i / (\|\mathbf{q}\|\|\mathbf{d}_i\|)$,
where $\mathbf{q}$ is the query embedding and $\mathbf{d}_i$ is the embedding of the $i$th patient-record chunk. The top-$k$ chunks are then concatenated into the LLM context window. This approach works well when the answer is contained in one or a few semantically similar passages. However, suicide risk formulation often requires integrating evidence distributed across diagnoses, medications, social history, life events, and longitudinal clinical transitions. These facts may be clinically related even when they are not lexically or semantically close in the original record.

GraphRAG addresses this limitation by making relationships explicit before retrieval. Instead of treating the patient record only as flat text, BEACON-SP maps structured patient attributes and derived constructs into a typed knowledge graph. Nodes represent entities such as diagnoses, medications, encounters, risk factors, protective factors, interventions, suicide-behavior constructs, and temporal events. Edges represent meaningful relationships such as diagnosis history, medication exposure, co-occurrence, temporal ordering, risk-factor membership, and ontology alignment. At query time, BEACON-SP uses semantic retrieval to identify an initial set of graph seeds and then expands from those seeds over typed graph edges. This graph traversal supplies the relational and temporal context needed to answer multi-hop clinical questions.

\begin{figure}[t]
   \centering
   \includegraphics[width=\linewidth,height=1.2in]{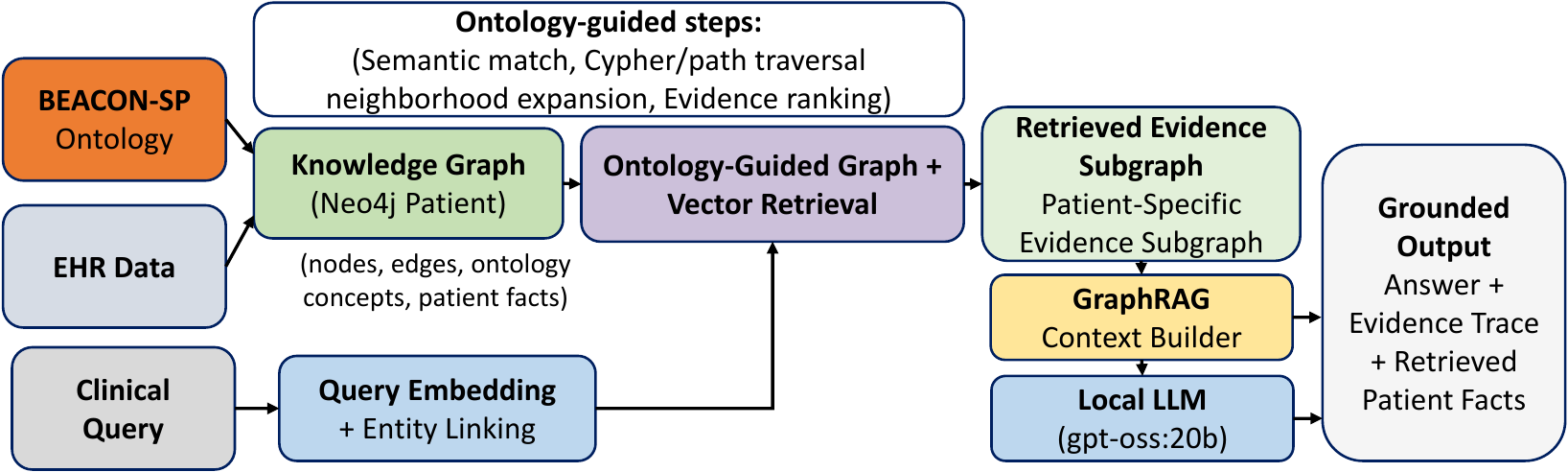}
   \caption{BEACON-SP GraphRAG system architecture.}
   \label{fig:GraphRAG_Architect}
\end{figure}

Figure~\ref{fig:GraphRAG_Architect} presents our framework that uses the BEACON-SP ontology as the schema to construct patient-level knowledge graphs rather than embedding documents as flat text. Structured patient data are mapped into graph nodes and edges according to the ontology. Suicide theory constructs  along with temporal features are added as higher-order nodes that capture causal and temporal relations among clinical events. The resulting graph is stored in Neo4j with patient-specific identifiers to prevent cross-patient contamination. A patient-specific subgraph is retrieved at query time and passed as structured context to the language model.

Operationally, BEACON-SP uses a two-pass retrieval strategy. First, the query is embedded and matched to patient-specific graph node descriptions using cosine similarity to select relevant seed nodes. Second, a bounded breadth-first expansion retrieves neighboring nodes and relations within the same patient graph, constrained by patient identifier, edge type, and traversal depth. The resulting subgraph is serialized as structured evidence with node labels, edge types, timestamps, and source attributes. This combines semantic retrieval with ontology-guided graph traversal, enabling BEACON-SP to support both direct questions, such as medication history, and relational questions about how diagnoses, stressors, recent events, and protective factors shape the risk profile of a patient.

% Evaluations

\section{Evaluations}\label{sec:experiments}

\subsection{Experimental Setup}
In our evaluations, we used the patient records in the Veterans Affairs Suicide Prevention Synthetic Dataset (VASPSD) dataset \cite{VASPSD} to create the BEACON-SP knowledge graph. We perform both qualitative and quantitative assessments. 

Our \textbf{qualitative assessments} focus on the responses of the two retrieval systems using the same query. To isolate the effect of retrieval structure, both systems use identical answer-generation models and system prompts, and differ only in whether context is described as coming from a vector store or a knowledge graph. 

For \textbf{quantitative assessments}, we compare the ontology-grounded GraphRAG against a vector-based RAG baseline on a benchmark of 1,500 clinical queries drawn from 100 patients across 15 suicide-prevention query categories: diagnosis, medication, suicide indicators, stressors, summary generation, comorbidity, temporal reasoning, protective factors, risk factors, substance interaction, ontology mapping, risk construct inference, risk window identification, protective versus risk balance, and lethal means assessment (100 queries per category). Both systems share the same answer-generation model and prompting and differ only in their retrieval mechanism. RAG retrieves semantically similar patient-record chunks from a vector store, whereas BEACON-SP GraphRAG retrieves patient-specific nodes and relationships from the knowledge graph using a two-pass strategy: a cosine-similarity seed pass followed by breadth-first graph expansion. This process surfaces structured relations among diagnoses, medications, social determinants, risk factors, protective factors, suicide-theory constructs, and temporal events, thereby isolating the effect of retrieval structure.

\begin{figure}[t!]
   \centering
   \begin{labeledbox}{questionblue}{questionbg}{Question}
   \scriptsize
       What is the temporal sequence of this patient's psychiatric history, including onset dates, episode progression, and key clinical transitions?
   \end{labeledbox}

   \begin{labeledbox}{graphorange}{graphbg}{GraphRAG}
   \scriptsize
   \centering
   \textbf{Chronology of Psychiatric Illness}
   \begin{minipage}{\textwidth}
   % \centering
   \raggedright
   \begin{tabular}{lccc}
   \toprule
   \textbf{Date} & \textbf{Diagnostic Event} & \textbf{Clinical Note / Episode}  \\
   \midrule
   1993-04-12   &   Major Depressive &  First documented MDD onset.\\&Disorder (MDD)& No psychotic features noted. \\
   &...&\\
    \multicolumn{3}{l}{Detailed EHR logs are omitted here due to space constraints.}   \\
    &...&\\
   2021-02-08 & Psychotic Depression &  Further episode recorded; \\
   &– additional onset & no remission documented.\\
   \bottomrule
   \end{tabular}
   \end{minipage}
   \\
   \vspace{2pt}
   \textbf{Episode Progression \& Key Transitions}

\begin{enumerate}[leftmargin=*, itemsep=0pt]
   \item \textbf{Initial MDD (1993)} – First major depressive disorder episode without psychosis.
   \item \textbf{First Psychotic Episode (2003)} – Depression acquires psychotic features, marking a more severe phenotype.
   \item \textbf{Emergence of Schizophrenia (2005)} – New schizophrenia diagnosis; antipsychotic therapy initiated.
   \item \textbf{Onset of Anxiety (2007)} – Anxiety disorder added to clinical picture.
   \item \textbf{Formal Psychotic-Depression Diagnosis (2013)} – Psychotic-depressive episodes formally coded as ``Psychotic Depression.''
   \item \textbf{Relapses (2016--2020)} – Recurrent psychotic-depression episodes indicating chronic, relapsing illness.
   \item \textbf{Co-morbid Schizophrenia \& Anxiety} – Persistent diagnoses throughout 2005--2020.
   \item \textbf{Last Documented Episode (2020--2021)} – Ongoing psychotic-depression with no remission recorded.
\end{enumerate}

\textbf{Clinical Take-away}

\begin{itemize}[leftmargin=*, itemsep=0pt]
   \item Long-standing recurrent MDD progressing to psychotic depression since 2003.
   \item Schizophrenia (2005) and anxiety (2007) add complexity to the psychiatric profile.
   \item Psychotic-depression becomes a chronic, distinct diagnosis from 2013 onward.
   \item Persistent, relapsing psychotic-depression with no documented remission through 2021.
   \item Ongoing management with long-term antidepressant and antipsychotic therapy.
\end{itemize}
   \end{labeledbox}

   \begin{labeledbox}{tradpurple}{tradbg}{Vector-based RAG}
   \scriptsize
   I have no retrieved patient data available to outline the temporal sequence of this patient's psychiatric history.
   \end{labeledbox}
   \caption{GraphRAG and Vector-based RAG responses to psychiatric history and progression.
   }
   \label{Fig:Mario_Ferrer_Social_RAG}
   \vspace{-0.1cm}
\end{figure}

\subsection{Qualitative Evaluations}

Figure~\ref{Fig:Mario_Ferrer_Social_RAG} illustrates the qualitative difference between GraphRAG and RAG using a representative multi-hop query about the psychiatric history of a high-risk patient from the VASPSD with PTSD, Generalized Anxiety Disorder (GAD), schizophrenia, and chronic pain.

When asked to reconstruct the patient's psychiatric history, GraphRAG returns a structured chronology anchored in discrete diagnostic events, including Major Depressive Disorder (MDD) onset in 1993, psychotic features in 2003, schizophrenia in 2005, anxiety in 2007, psychotic depression in 2013, and a relapsing course through 2021 with no documented remission. The response organizes these events into a dated sequence and summarizes the trajectory in terms of chronicity, comorbidity, and ongoing pharmacologic management.

RAG returns no relevant information for the same query. Although the relevant facts are present in the underlying record, vector retrieval fails to assemble the longitudinal diagnostic evidence to answer a temporally structured question.

This example highlights the advantage of ontology-grounded retrieval. Reconstructing psychiatric history requires linking diagnoses across time, distinguishing onset from relapse, and tracing accumulating comorbidity. These relations are explicitly encoded in the graph and recoverable through traversal, whereas flat chunk retrieval ranks locally similar text without a mechanism for assembling the full clinical trajectory.

\subsection{Quantitative Evaluations}

We evaluated the system responses using an LLM-as-judge protocol. Each response is scored on four criteria using a trinary scale (0, 0.5, 1):

\begin{itemize}
    \item \textbf{Factual Accuracy:} whether clinical claims are correct with respect to the patient record.
    \item \textbf{Completeness:} whether all relevant evidence in the patient record is captured.
    \item \textbf{Clinical Relevance:} whether the answer interprets evidence clinically rather than listing raw facts.
    \item \textbf{Evidence Grounding:} whether specific claims are traceable to patient-record evidence.
\end{itemize}

We employ two evaluation modes. In \textit{absolute} mode, each system is scored independently against the structured patient record. In \textit{comparative} mode, the judge receives both answers simultaneously alongside each system's retrieved evidence (text chunks for RAG, graph nodes for GraphRAG), and scores both before declaring a winner per criterion.

The distinction carries methodological weight. Under absolute scoring, graph nodes retrieved by GraphRAG cannot be verified against the flat patient record, so ontological constructs such as risk windows and theoretical framework nodes are penalised as ungrounded even when correctly retrieved. The comparative judge corrects this by treating each system's retrieved evidence as the appropriate verification source: RAG answers are checked against the patient record and retrieved chunks, GraphRAG answers against the patient record \textit{and} retrieved graph nodes. This correction is necessary for a fair assessment of ontology-grounded retrieval and constitutes a methodological contribution of this work.

\begin{table}[tb!]
\centering
\caption{Overall mean absolute judge scores (1,500 queries, 100 patients, 15 tasks). Best score per criterion in bold.}
\label{tab:overall_scores}
\renewcommand{\arraystretch}{1.25}
\setlength{\tabcolsep}{3.2pt}
\begin{tabular}{lcccc}
\toprule
\textbf{System} &
\begin{tabular}[c]{@{}c@{}}\textbf{Factual}\\\textbf{Accuracy}\end{tabular} &
\begin{tabular}[c]{@{}c@{}}\textbf{Comple-}\\\textbf{teness}\end{tabular} &
\begin{tabular}[c]{@{}c@{}}\textbf{Clinical}\\\textbf{Relevance}\end{tabular} &
\begin{tabular}[c]{@{}c@{}}\textbf{Evidence}\\\textbf{Grounding}\end{tabular} \\
\midrule
RAG      & \textbf{0.424} & 0.348 & 0.193 & \textbf{0.432} \\
GraphRAG & 0.231          & \textbf{0.342} & \textbf{0.359} & 0.200 \\
\bottomrule
\end{tabular}
\end{table}

\begin{table}[tb!]
\centering
\caption{Paired comparative evaluation (1,500 queries $\times$ 4 criteria, minus skipped pairs).}
\label{tab:paired_scores}
\renewcommand{\arraystretch}{1.2}
\begin{tabular}{lc}
\toprule
\textbf{Outcome} & \textbf{Count / Rate} \\
\midrule
GraphRAG wins      & 4,482 \\
RAG wins           & 238 \\
Ties               & 1,144 \\
Skipped            & 136 \\
Total comparisons  & 5,864 \\
GraphRAG win rate  & 76.4\% \\
RAG win rate       &  4.1\% \\
Tie rate           & 19.5\% \\
\bottomrule
\end{tabular}
\end{table}

\begin{table}[t!]
\centering
\caption{GraphRAG win rate by criterion under the comparative judge.}
\label{tab:criterion_win_rates}
\resizebox{\columnwidth}{!}{%
\begin{tabular}{lcccc}
\toprule
\textbf{Criterion} & \textbf{GR Wins} & \textbf{RAG Wins} & \textbf{Ties} & \textbf{GR Win Rate} \\
\midrule
Factual Accuracy   &   716 &  70 & 680 & 48.8\% \\
Completeness       & 1,231 & 100 & 135 & 84.0\% \\
Clinical Relevance & 1,377 &  23 &  66 & 93.9\% \\
Evidence Grounding & 1,158 &  45 & 263 & 79.0\% \\
\bottomrule
\end{tabular}}
\end{table}

\begin{table*}[t!]
\centering
\caption{Mean absolute judge scores per task (100 patients). Best score per RAG--GraphRAG pair bolded.}
\label{tab:per_task_scores}
\renewcommand{\arraystretch}{1.15}
\setlength{\tabcolsep}{4.2pt}
\begin{tabular}{lcccccccc}
\toprule
& \multicolumn{4}{c}{\textbf{RAG}} & \multicolumn{4}{c}{\textbf{GraphRAG}} \\
\cmidrule(lr){2-5} \cmidrule(lr){6-9}
\textbf{Task} &
\begin{tabular}[c]{@{}c@{}}\textbf{Factual}\\\textbf{Accuracy}\end{tabular} &
\begin{tabular}[c]{@{}c@{}}\textbf{Comple-}\\\textbf{teness}\end{tabular} &
\begin{tabular}[c]{@{}c@{}}\textbf{Clinical}\\\textbf{Relevance}\end{tabular} &
\begin{tabular}[c]{@{}c@{}}\textbf{Evidence}\\\textbf{Grounding}\end{tabular} &
\begin{tabular}[c]{@{}c@{}}\textbf{Factual}\\\textbf{Accuracy}\end{tabular} &
\begin{tabular}[c]{@{}c@{}}\textbf{Comple-}\\\textbf{teness}\end{tabular} &
\begin{tabular}[c]{@{}c@{}}\textbf{Clinical}\\\textbf{Relevance}\end{tabular} &
\begin{tabular}[c]{@{}c@{}}\textbf{Evidence}\\\textbf{Grounding}\end{tabular} \\
\midrule
Comorbidity              & 0.060 & 0.070 & 0.000 & 0.035 & \textbf{0.260} & \textbf{0.260} & \textbf{0.485} & \textbf{0.195} \\
Diagnosis                & 0.315 & 0.310 & 0.320 & 0.280 & \textbf{0.140} & \textbf{0.305} & \textbf{0.360} & \textbf{0.135} \\
Lethal Means Assessment  & \textbf{0.625} & \textbf{0.620} & 0.145 & \textbf{0.520} & 0.425 & 0.390 & \textbf{0.220} & 0.255 \\
Medication               & \textbf{0.580} & \textbf{0.620} & 0.565 & \textbf{0.640} & 0.240 & 0.390 & \textbf{0.495} & 0.210 \\
Ontology                 & 0.000 & 0.045 & 0.000 & 0.000 & \textbf{0.030} & \textbf{0.155} & \textbf{0.005} & \textbf{0.010} \\
Protective Factors       & \textbf{0.825} & 0.405 & 0.025 & \textbf{0.785} & 0.210 & \textbf{0.375} & \textbf{0.115} & 0.155 \\
Protective vs.\ Risk     & \textbf{0.315} & 0.270 & 0.185 & \textbf{0.400} & 0.105 & \textbf{0.270} & \textbf{0.450} & 0.185 \\
Risk Construct Inference & \textbf{0.465} & \textbf{0.405} & 0.030 & \textbf{0.370} & 0.310 & 0.345 & \textbf{0.460} & 0.270 \\
Risk Factors             & 0.065 & 0.140 & 0.010 & 0.095 & \textbf{0.165} & \textbf{0.360} & \textbf{0.275} & \textbf{0.145} \\
Risk Window Identification & \textbf{0.690} & \textbf{0.385} & 0.110 & \textbf{0.530} & 0.255 & 0.285 & \textbf{0.515} & 0.185 \\
Stressor                 & \textbf{0.475} & \textbf{0.440} & \textbf{0.375} & \textbf{0.770} & 0.270 & 0.470 & 0.250 & 0.400 \\
Substance Interaction    & \textbf{0.650} & 0.320 & 0.055 & \textbf{0.700} & 0.295 & \textbf{0.380} & \textbf{0.360} & 0.195 \\
Suicide Indicators       & \textbf{0.795} & \textbf{0.810} & \textbf{0.840} & \textbf{0.800} & 0.465 & 0.525 & 0.530 & 0.405 \\
Summary                  & \textbf{0.270} & 0.150 & 0.185 & \textbf{0.295} & 0.175 & \textbf{0.405} & \textbf{0.410} & 0.120 \\
Temporal                 & \textbf{0.230} & 0.230 & 0.045 & \textbf{0.260} & 0.125 & \textbf{0.215} & \textbf{0.450} & 0.130 \\
\midrule
\textbf{Overall}         & \textbf{0.424} & 0.348 & 0.193 & \textbf{0.432} & 0.231 & \textbf{0.342} & \textbf{0.359} & 0.200 \\
\bottomrule
\end{tabular}
\end{table*}

Table~\ref{tab:overall_scores} reports absolute scores. The systems split by criterion: RAG leads on factual accuracy (0.424 vs.\ 0.231) and evidence grounding (0.432 vs.\ 0.200), while GraphRAG leads on clinical relevance (0.359 vs.\ 0.193), with near parity on completeness. This reflects a structural asymmetry in the absolute judge: claims are verified against the flat patient record, so RAG answers derived from the same text fields receive higher grounding scores, whereas GraphRAG answers citing knowledge-graph constructs absent from the flat record are penalized as ungrounded. Thus, the absolute results conflate proximity to the source representation with answer quality.

% The comparative judge corrects this asymmetry. 
The comparative judge mitigates this representation asymmetry. Across 5,864 criterion-level comparisons in Table~\ref{tab:paired_scores}, GraphRAG is preferred in 4,482 cases (76.4\%), compared with 238 cases (4.1\%) for RAG and 19.5\% ties. Under this fairer protocol, ontology-grounded retrieval produces consistently stronger clinical answers. Table~\ref{tab:criterion_win_rates} shows that GraphRAG wins 93.9\% of clinical relevance and 84.0\% of completeness comparisons, reflecting the ontology's ability to surface structured relational evidence in clinically interpretable form. Evidence grounding reaches 79.0\%, reversing the absolute judge's finding and confirming that RAG's earlier advantage was an attribution artifact. Factual accuracy shows the smallest comparative advantage: GraphRAG wins 48.8\% of comparisons, while 46.4\% are ties and 4.8\% favor RAG.
%Factual accuracy is nearly tied at 48.8\%, indicating that neither system is meaningfully more accurate when evaluated against its own evidence.

%The comparative judge corrects this asymmetry. Across 5,864 criterion-level comparisons shown in Table~\ref{tab:paired_scores}, GraphRAG is preferred in 4,482 cases (76.4\%) and RAG in only 238 (4.1\%), with 19.5\% ties. Under a fair evaluation protocol, ontology-grounded retrieval produces consistently superior clinical answers. Table~\ref{tab:criterion_win_rates} breaks results down by criterion. GraphRAG wins 93.9\% of clinical relevance comparisons and 84.0\% of completeness comparisons, reflecting the ontology's ability to surface structured relational evidence in clinically interpretable form. Evidence grounding reaches 79.0\%, a reversal of the absolute judge's finding, which confirms the earlier apparent advantage of RAG was an attribution artifact. Factual accuracy is nearly tied at 48.8\%, indicating that neither system is meaningfully more accurate when each is evaluated against its own evidence.

Table~\ref{tab:per_task_scores} reports the task-level absolute scores and shows that the two retrieval strategies have different strengths. RAG generally achieves higher factual accuracy and evidence grounding, while GraphRAG achieves higher clinical relevance in 12 of 15 tasks, trailing only on medication, stressor, and suicide indicators. The clearest exception is the suicide-indicators task, where RAG obtains the highest scores across all four criteria (0.795, 0.810, 0.840, and 0.800). This reflects the structure of the VASPSD records, where suicide-related indicators often appear in compact clinical entries that vector retrieval can recover intact. In contrast, GraphRAG distributes the same information across multiple node types, so missing any relevant node can reduce coverage. This is the one category where flat text retrieval retains a genuine advantage. Ontology mapping and protective-versus-risk balance remain challenging for both systems, suggesting that these tasks require stronger reasoning over theoretical constructs and competing clinical evidence beyond retrieval alone.

%For the task-level results, Table~\ref{tab:per_task_scores} presents the task-level absolute scores. RAG tends to score higher on factual accuracy and evidence grounding, while GraphRAG leads on clinical relevance in 12 of 15 tasks, trailing on medication, stressor, and suicide indicators. The clearest case is suicide indicators, where RAG scores highest across all four criteria (0.795, 0.810, 0.840, 0.800). Suicide-related documentation in the VASPSD often appears as a single dense clinical entry that RAG retrieves intact; GraphRAG distributes the same information across multiple node types, and incomplete retrieval of any one degrades coverage. This is the one category where flat text retrieval retains a genuine advantage. Ontology mapping and protective versus risk balance are challenging for both systems, reflecting the limited presence of theoretical framework nodes in the flat record and the difficulty of generating clinically reasoned balance assessments from retrieval alone.

\begin{table}[t!]
\centering
\caption{GraphRAG paired win rate by task using the comparative judge (100 patients).}
\label{tab:paired_task_win_rates}
\renewcommand{\arraystretch}{1.1}
\begin{tabular}{lc}
\toprule
\textbf{Task} & \textbf{GraphRAG Win Rate} \\
\midrule
Risk Factors                 & 91.5\% \\
Ontology                     & 89.4\% \\
Comorbidity                  & 85.1\% \\
Risk Construct Inference     & 84.7\% \\
Protective vs.\ Risk Balance & 84.5\% \\
Lethal Means Assessment      & 83.0\% \\
Summary                      & 82.1\% \\
Diagnosis                    & 78.8\% \\
Temporal                     & 78.5\% \\
Risk Window Identification   & 76.3\% \\
Substance Interaction        & 74.2\% \\
Stressor                     & 64.8\% \\
Medication                   & 62.0\% \\
Protective Factors           & 61.0\% \\
Suicide Indicators           & 51.3\% \\
\midrule
\textbf{Overall}             & \textbf{76.4\%} \\
\bottomrule
\end{tabular}
\end{table}

For paired task-level comparison, Table~\ref{tab:paired_task_win_rates} reports GraphRAG win rates per task. Win rates exceed 80\% on seven tasks: risk factors (91.5\%), ontology (89.4\%), comorbidity (85.1\%), risk construct inference (84.7\%), protective versus risk balance (84.5\%), lethal means assessment (83.0\%), and summary (82.1\%). Each of these requires multi-hop reasoning across heterogeneous node types: integrating diagnoses with theoretical suicide constructs, which links conditions to temporal risk windows, or synthesizing competing risk and protective evidence. The BEACON-SP ontology encodes these relations explicitly; vector similarity cannot reconstruct them from chunk co-occurrence.

In areas where RAG is competitive such as suicide indicators (51.3\%), protective factors (61.0\%), medication (62.0\%), and stressor identification (64.8\%), the relevant clinical information is largely self-contained in a small number of structured EHR fields and retrieved accurately by both systems. For such tasks, the relational scaffolding of the knowledge graph adds less marginal value.

\subsection{Discussion}

Two findings emerge from our evaluations. The first finding is methodological. When claims are verified only against flat patient records, GraphRAG appears weaker in factual accuracy and evidence grounding. This is partly a measurement artifact: graph-derived constructs may not appear explicitly in the source text. The comparative judge reduces this asymmetry by evaluating each system against its own retrieved evidence. Under this protocol, GraphRAG achieves a 79.0\% evidence-grounding win rate, while factual accuracy is essentially tied at 48.8\%, suggesting that flat-text ground truth can underestimate graph-based retrieval.

The second finding is clinical. After correcting evidence attribution, BEACON-SP GraphRAG wins 76.4\% of 5,864 head-to-head criterion evaluations, with the largest gains in clinical relevance (93.9\%) and completeness (84.0\%). These improvements are concentrated in relational reasoning tasks, including risk factor identification, comorbidity analysis, risk construct inference, protective-versus-risk balance, and summary synthesis. Such tasks benefit from the explicit structure encoded in the BEACON-SP ontology, which flat passage retrieval cannot recover from semantic similarity alone.

Suicide indicators are the main exception. RAG performs better under both protocols, likely because dense clinical-note passages capture these indicators more completely than distributed graph nodes. Similar patterns for protective factors, medications, and stressors motivate a hybrid design in which graph retrieval supports relational inference, while passage retrieval supplies dense clinical documentation. Both systems remain below 0.5 on protective-versus-risk balance, ontology mapping, and summary tasks, indicating that high-level synthesis requires stronger reasoning and grounding beyond retrieval architecture alone.

\section{Conclusion}\label{sec:discussion}

This paper presented BEACON-SP, an ontology-grounded GraphRAG framework for clinician-facing suicide-prevention decision support. BEACON-SP combines patient knowledge graphs with ontology-guided retrieval to provide structured reasoning over clinical, behavioral, social, and temporal evidence. The framework uses a comprehensive suicide prevention ontology that integrates three suicide-prevention theories into a unified representation of diagnoses, medications, medical conditions, risk and protective factors, interventions, life events, suicide-behavior constructs, and temporal risk windows. We evaluated BEACON-SP for clinician-facing question answering using an ontology-guided GraphRAG workflow. On a benchmark of 1,500 queries spanning 15 clinical categories and 100 patients, GraphRAG was preferred in 76.4\% of head-to-head comparisons, with gains in clinical relevance (93.9\%) and completeness (84.0\%). We also identified and corrected an evaluation bias that penalizes graph-based systems for citing ontological constructs absent from flat EHR records. The resulting comparative protocol provides a more equitable assessment of ontology-grounded retrieval systems. BEACON-SP establishes a foundation for interpretable, evidence-grounded clinical decision support. Future work will focus on larger benchmarks, clinician review, and hybrid graph-text retrieval.
 
\bibliographystyle{ieeetr}
\bibliography{references}

\end{document}